\documentclass{article}

  \usepackage[final]{nips_2016}
  \usepackage{natbib}
  \setcitestyle{numbers}
  \usepackage[utf8]{inputenc} 
  \usepackage[T1]{fontenc}    
  \usepackage{hyperref}       
  \usepackage{url}            
  \usepackage{booktabs}       
  \usepackage{amsfonts}       
  \usepackage{nicefrac}       
  \usepackage{microtype}      
  \usepackage{indentfirst}
  \usepackage{amsmath}
  \usepackage{fancyhdr}
  \usepackage{graphicx}
  \usepackage{printlen}
  \usepackage{color}

\fancypagestyle{equalc}{\fancyhf{}\fancyfoot[R]{* indicates equal contribution}}

\title{A Comparison of LSTMs and Attention Mechanisms for Forecasting Financial Time Series}

\author{
    S.E. Yi* \\
    Department of Computer Science\\
    University of Toronto\\
    Toronto, ON M5S 3H7 \\
    \texttt{seungeunyi@cs.toronto.edu} \\
    \And
    A. Viscardi* \\
    Department of Computer Science\\
    University of Toronto\\
    Toronto, ON M5S 3H7 \\
    \texttt{avis@cs.toronto.edu} \\
    \And
    T. Hollis* \\
    Department of Computer Science\\
    University of Toronto\\
    Toronto, ON M5S 3H7 \\
    \texttt{thollis@cs.toronto.edu} \\
}

\begin{document}

\maketitle

\begin{abstract}

  While LSTMs show increasingly promising results for forecasting Financial Time Series (FTS), this paper seeks to assess if attention mechanisms can further improve performance. The hypothesis is that attention can help prevent long-term dependencies experienced by LSTM models. To test this hypothesis, the main contribution of this paper is the implementation of an LSTM with attention. Both the benchmark LSTM and the LSTM with attention were compared and both achieved reasonable performances of up to 60\% on five stocks from Kaggle's Two Sigma dataset. This comparative analysis demonstrates that an LSTM with attention can indeed outperform standalone LSTMs but further investigation is required as issues do arise with such model architectures.  
  
\end{abstract}

\thispagestyle{equalc}

\section{Introduction}

Financial Time Series (FTS) modelling is a practice with a long history which first revolutionised algorithmic trading in the early 1970s. The analysis of FTS was divided into two categories: fundamental analysis and technical analysis. Fundamental analysis is the study of a stock or currency’s price based on economic factors. These include the overall state of the business or economy, revenue, interest rates and others. On the other hand, technical analysis, as defined by J. Murphy in \cite{murphy1999technical}, is the study of market action to forecast future trends. This is achieved through the analysis of shorter term financial data, primarily price and volume. Both fundamental and technical analysis are put into question by the Efficient Market Hypothesis (EMH). The EMH, highly disputed since its initial publication in 1970, hypothesizes that stock prices are ultimately unpredictable \cite{malkiel1970efficient}. This has not stopped research attempting to model FTS through the use of linear, non-linear and ML-based models, as mentioned hereafter. Together these approaches form the main subcategories of existing solutions in FTS analysis. 

In parallel, the origins of attention mechanisms initially came into prominence in the field of Computer Vision (CV). These new models were originally loosely inspired by human visual attention mechanisms present in a region of the prefrontal cortex known as the inferior frontal junction. Applications were later leveraged to tackle issues with context-dependent inference in Natural Language Processing (NLP).  In both cases, the core principle of attention is achieved by setting attention weights to assign more or less of the algorithm’s finite attention on different subsets of the input feature space. In CV, this corresponds to focussing on particular features of the input image while in NLP this represents focus on particular words in the input sentence. In NLP, this attention mechanism allows inference to be made about an entire sentence while remaining sensitive to its context. This remains to this day a particularly challenging task for long sentences.

In this paper we propose a novel approach to FTS forecasting by combining these two fields of research. The idea is to leverage the developments in attention mechanisms to improve the performance of promising LSTM RNN architectures currently in use for FTS forecasting. The main contribution in this paper is the implementation of an LSTM that uses attention for parsing both news headlines and financial data. The performance of this model is then compared with that of a regular LSTM without attention. This performance is then evaluated using five stocks from Kaggle's Two Sigma dataset \cite{kaggle2017twosigma} and using various methods of data preprocessing \cite{bergmeir2012use}. 

The ultimate goals of FTS forecasting, among others, is to help solve the problem of volatility in speculative markets and to help foresee large financial events such as the 2009 financial crisis to ensure better economic preparation. 

\section{Related Work}

As discussed in \cite{hollis2018deep}, the most rudimentary approach to modelling FTS is by assuming that they follow the random walk model. The random walk model can be simply expressed as a sum of a series of independent random variables \cite{hamilton1994time}. By weighing these, the first Auto-Regressive (AR) model was developed. A set of equations were developed by U. Yule and G. Walker in \cite{walker1931periodicity} to provide quantitative methods for estimating parameters in AR models. This work was subsequently expanded upon by J. P. Burg in \cite{burg1968new} who provided an alternative approach albeit with different stability properties. These AR models are often accompanied by another type of linear model, the Moving Average (MA) model which gives the Auto-Regressive Moving Average (ARMA) model. However, a fundamental limitation of AR, MA and ARMA models is that they all assume the process being modelled is stationary. Stationarity is a property of processes whereby the probability distribution remains constant over time, thus variance also remains constant. Indeed, this assumption is significant as FTS are often non-stationary processes. Therefore, this model’s accuracy will suffer, highlighting the need to take this problem of stationarity into consideration. This is done by generalising the ARMA model into the Autoregressive Integrated Moving Average (ARIMA) model \cite{hamilton1994time}. The ARIMA model solves the issue of non-stationarity by exploiting the concept of returns (or degrees of differencing). Non-stationary time series can therefore be made stationary by differencing. The aforementioned linear models all suffer from the assumption that FTS are homoscedastic processes. This is indeed often a poor assumption to make, as shown in \cite{engle1982autoregressive} by R.F. Engle. In \cite{engle1982autoregressive}, Engle states that by using a more sophisticated model such as the Auto-Regressive Conditional Heteroscedasticity (ARCH) model, the homoscedastic assumption can be avoided. This ARCH model was later described by Bollerslev in \cite{bollerslev1986generalized} as a special case of a more generalised model called the Generalised Auto-Regressive Conditional Heteroscedasticity (GARCH) model. Many more variants of the GARCH model have been published since its original publication in 1986. These include NAGARCH (nonlinear asymmetric GARCH) \cite{engle1993measuring}, EGARCH (exponential GARCH) \cite{pierre1998estimating}, GJR-GARCH (Glosten-Jagannathan-Runkle GARCH) \cite{hamzaoui2016glosten} and many others. These GARCH derivatives are often nested under Hentschel’s fGARCH (Family GARCH) model \cite{hentschel1995all} but these all lie outside the scope of this paper. In the same time as the ARCH and GARCH models, J. Leontaris and S. A. Billings published an alternative in \cite{chen1989representations} known as is the Nonlinear Autoregressive Moving Average model with exogenous inputs (NARMAX). This work, building on their own previous work on ARMAX models, demonstrated that NARMAX models can successfully be applied to model complex time series. More information on these models can be found in \cite{hollis2018deep}, including equations and further explanation. 
				
These state space and stochastic models were however quickly overwhelmed by advances in Machine Learning. A wave of ML approaches to modelling FTS severely disrupted the field of algorithmic trading via stochastic modelling in the last two decades. One of the earliest approaches to FTS forecasting using ML built on work from Kohonen in \cite{kohonen1982self}. In \cite{kohonen1982self}, Kohonen introduced the idea of Self-Organising Maps (SOM) which were subsequently successfully applied to FTS forecasting \cite{koskela1998time}. In 2003, still in the early days of ML for FTS predictions, SVMs (both linear and non-linear) were shown by Kim in \cite{kim2003financial} to be of significant  predictive capabilities for FTS. In parallel, Zhang showed in his 2003 paper \cite{zhang2003time} that, by combining Artificial Neural Networks (ANNs) with the aforementioned ARIMA model, promising FTS forecasting can be achieved. 

Nevertheless, it was only by benefitting from the Neural Network boom of 2012 brought on by the AlexNet work of Krizhevsky, Sutskever and Hinton on the ImageNet competition \cite{krizhevsky2012imagenet}, that ANNs became some of the most mainstream methods for FTS forecasting, in particular with the rise of RNNs \cite{chandra2012cooperative}. However, another type of neural network that has also been widely lauded for its performance in FTS forecasting is the Echo State Network (ESN). Indeed, Lin et. al showed in \cite{lin2009short} that ESNs combined with Principal Component Analysis (PCA) can sometimes exceed or at least match the performance of conventional RNNs while decreasing the computational costs. This is due to the techniques of Reservoir Computing introduced in ESNs. In short, ESNs can bypass the issue of the vanishing gradient problem and long computational times, present in conventional RNNs, by creating a large `reservoir' of sparsely connected neurons. The connections are assigned at random and weights within the reservoir do not get conventionally trained, reducing computational time and allowing the network to echo past states (emulating the `memory' of RNNs).  

Another alternative to RNNs is the Deep Belief Network (DBN). Hinton and Salakhutdinov's DBNs are a type of probabilistic generative neural network composed of layers of Restricted Boltzmann Machines (RBMs). These have also successfully been leveraged for accurate time series prediction \cite{kuremoto2014time}.

In the modern day however, LSTM RNNs remain amongst some of the most popular and promising models for predicting FTS. While LSTMs originally came to light in the seminal 1997 paper \cite{hochreiter1997long} by Hochreiter and Schmidhuber, they only recently rose to prominence for FTS forecasting. Amongst the most successful LSTM implementations, the pioneering paper by Bao et al. in \cite{bao2017deep} implements a variety of the most modern LSTM architectures coupled with autoencoders and applies them to stock data. The work presented here extends and builds on the insight of this paper by exploring the impact of leveraging attention models for sentiment analysis built on top of LSTMs. 

However, a word of caution is worth mentioning here. It is true that academic publications in the field of FTS forecasting are often misleading. Indeed, many of the most performant models are developed by private companies and kept away from the public, with the utmost secrecy for competitive reasons. Academia seems to be struggling to shed light on the most modern techniques which is one of the prime motivations for the investigation presented hereafter. In addition, many FTS forecasting papers tend to inflate their performance for recognition and overfit their models due to the heavy use simulators. Many of the performances claimed in these papers are difficult to replicate as they fail to generalise for future changes in the particular FTS being forecast.

Having reviewed the field of FTS forecasting, in order to better situate our paper amongst existing literature, it is important to now cover a brief history of attention mechanisms. 

Many early prominent papers using attention mechanisms in NLP initially used the term “alignment” to refer to attention in the context neural machine translation. One of the most foundational of these is the 2014 Bahdanau, Cho and Bengio collaboration in \cite{bahdanau2014neural}. In \cite{bahdanau2014neural}, Bahdanau shows that a single neural network can be jointly tuned to maximise translation performance using attention-based encoder-decoders. In parallel, Graves, Wayne and Danihelka showed in \cite{graves2014neural} that neural networks (in particular LSTMs) can be improved by coupling them to attention processes. While the application in \cite{graves2014neural} was helping Neural Turing Machines infer copying, sorting and recall, subsequent CV and NLP applications of the very same concept were greatly inspired by techniques presented in this paper \cite{xu2015show}. Building on the work above as well as on their own work, in 2015 Firat, Cho and Bengio in \cite{firat2016multi} had a breakthrough in multilingual Neural Machine Translation (NMT). They used single attention mechanisms shared across language pairs to outperform existing NMT benchmarks. 

Since then, a whole flurry of different types of attention have been developed including but not limited to self-attention, shared attention, local attention, hybrid attention and multi-headed attention. Indeed, the multi-headed attention model presented in the Transformer \cite{vaswani2017attention} by Google Brain and UofT alumni has been widely lauded as a promising novel architecture for dealing with long range dependencies without the issues of LSTMs in NMT. For this reason, that paper will serve as inspiration for our investigation on the potential of attention in FTS forecasting.

From the history of attention mechanisms presented here we can indeed see that attention mechanisms are currently accepted as a very promising approach to many problems, especially in the field of machine translation. However, we have yet to see if they present any promise in sentiment analysis for FTS forecasting which is the purpose of the work presented here. Intuitively, since attention mechanisms are designed to help address the issue of context dependency this suggests promising potential for sentiment analysis in FTS forecasting. Indeed, most sentiment analysis algorithms would fail to consider the impact of such context for long lengths. A naive example of this, such as “I am a compulsive liar but this company is fantastic, the stock is destined to rise and I don’t understand why more people have not invested yet”, would be wrongly detected as a positive sentiment by most algorithms. The following investigation aims to confirm or refute the hypothesis that LSTM performance can be improved with attention. 

\section{Model Architecture}

The two models investigated in this paper are a vanilla LSTM (as a benchmark) and an LSTM with an attention mechanism. A diagram of these models is shown in figure 1 below. 

\begin{figure}[!h]
	\includegraphics[width=375pt]{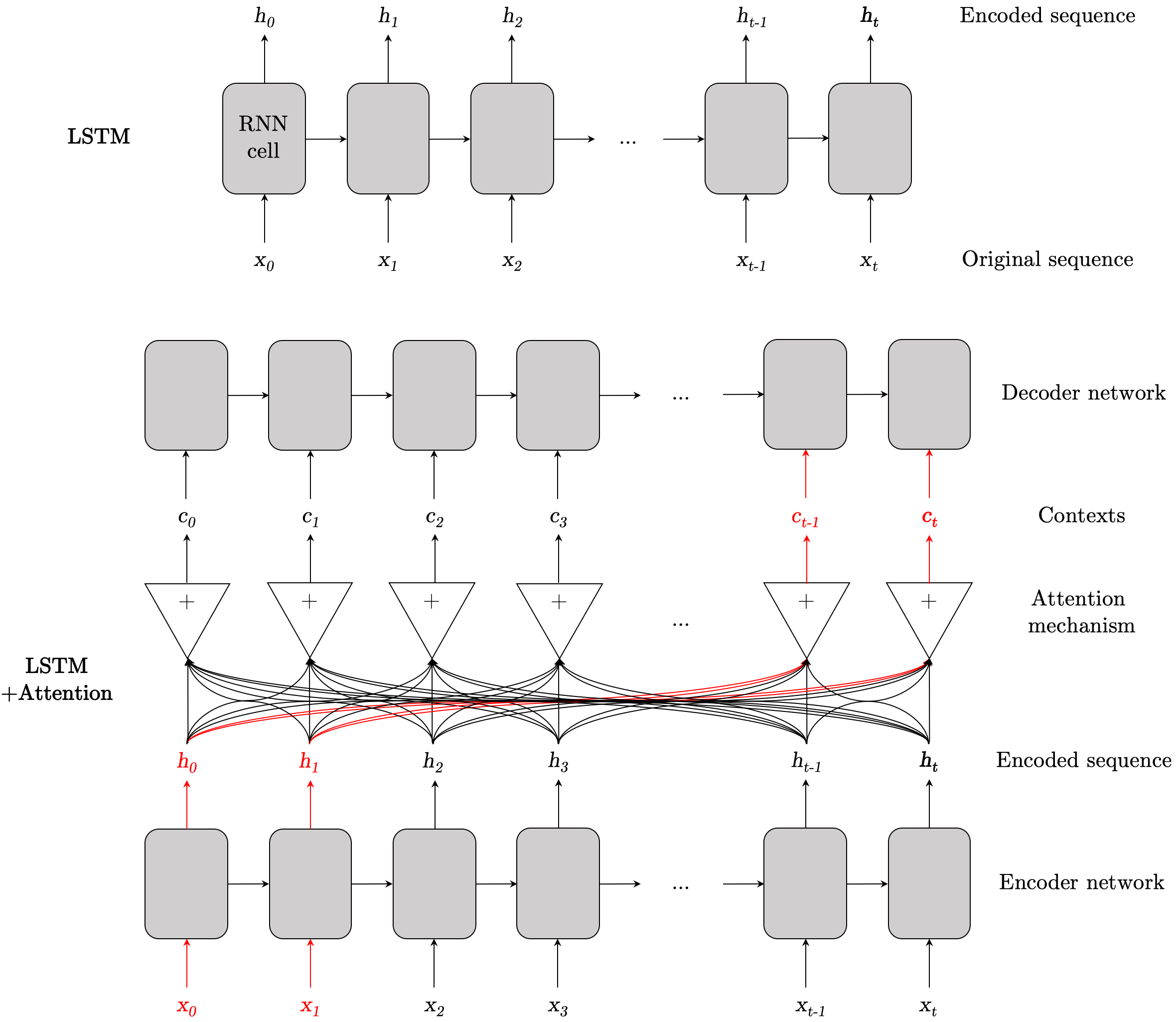}
	\caption{System diagram of the LSTM and LSTM with attention}
\end{figure}

Both the LSTM model and the LSTM with attention model used in this paper are implemented with mean squared error loss using an Adam optimiser \cite{kingma2014adam}. It is nonetheless important to cover the mathematical foundations of these models before comparing their performances.

In encoder-decoder RNNs, the encoder reads an input sequence $\textbf{x} = (x_1,$ ... $, x_{t-1})$ into a vector $\textbf{c} = (c_1,$ ... $, c_{t-1})$, as shown in figure 1. A common approach and the one we will be following in this paper is to use an RNN such that \cite{bahdanau2014neural}:
\begin{align}
    h_t = f(x_t, h_{t-1})
\end{align}
and
\begin{align}
    \textbf{c} = q(\{h_0, \dots , h_{T_x}\})
\end{align}
where $h_t$ is a hidden state at time $t$, $\textbf{c}$ is the context vector generated by the hidden states  and $f$ and $q$ are non-linear functions. 

Subsequently to equations 1 and 2, prediction is done at the decoder by defining a probability over the translation $\textbf{y}$ through the following decomposition \cite{bahdanau2014neural}:
\begin{align}
    p(\textbf{y}) = \prod_{t=1}^T p(y_t | \{y_1, \dots , y_{t-1}\}, c)
\end{align}
where $\textbf{y} = (y_1, \dots , y_{T_y})$. For RNNs, each conditional probability is modelled as:
\begin{align}
    p(y_t | \{y_1, \dots , y_{t-1}\}, c) = g(y_{t-1},s_t,c)
\end{align}
where $s_t$ is the RNN's hidden state and $g$ is a nonlinear function that outputs the probability of $y_t$. 

In attention, a particular context $c_i$ depends on a sequence of annotations ($h_1, \dots , h_{T_x}$) to which an encoder maps the input sequence. While each annotation $h_i$ contains information about the input sequence, we want to focus attention on a particular part of the input. Thus the context vector is computed as a weighted sum as follows \cite{bahdanau2014neural}:
\begin{align}
    c_i = \sum_{j=1}^{T_x} \alpha_{ij}h_j
\end{align}
where each context weight $\alpha_{ij}$ for each annotation $h_j$ is calculated as follows:
\begin{align}
    \alpha_{ij} = \frac{\exp(e_{ij})}{\sum_{k=1}^{T_x} \exp(e_{ik})}
\end{align}
where
\begin{align}
    e_{ik} = a(s_{i-1},h_j)
\end{align}
is known as the alignment model. Alignment models score how well matched the input at position $j$ and output at position $i$ are.

Equations 1 through 6 are used for the LSTM and attention mechanisms while Adam is used as the optimiser and Mean Square Error (MSE) is used as the loss function. Mathematically, MSE loss is as follows:
\begin{align}
    MSE = \frac{1}{n} \sum_{t=1}^{n} (y-t)^2
\end{align}

\section{Comparison of LSTM and LSTM with attention}

We can now move onto the implementation of the baseline LSTM. Both this baseline LSTM and the LSTM with attention were implemented within the Kaggle kernel environment in Python using the Keras library. It is worth noting here that special care was taken for sensible crosschecks. For example, the Kaggle kernel environment was set up to block the use of `future' data when training, preventing look-ahead bias. Indeed, look-ahead bias is a significant source of malpractice in the field of FTS so was worthy of extra consideration here. In addition, all implementations closely followed existing literature from the related work discussed in section 2.

To gain a good benchmark for the performance of an LSTM on forecasting stock prices from the Two Sigma dataset, we first consider a subset of stocks. This subset of stocks contains three very large companies (Intel, Wells Fargo, Amazon), one SME (Agilent Technologies) and one smaller company (Benchmark Electronics). These stocks were carefully chosen to have a wide variety of market cap, volatility and overall trend. Larger market cap stocks tend to be less volatile compared to smaller stocks. In addition, tech companies (like Amazon) tend to be less affected by the 2009 crash than finance companies (like Wells Fargo). These stocks and their volatility are shown in figure 2 as follows.
\begin{figure}[!h]
	\includegraphics[width=395pt]{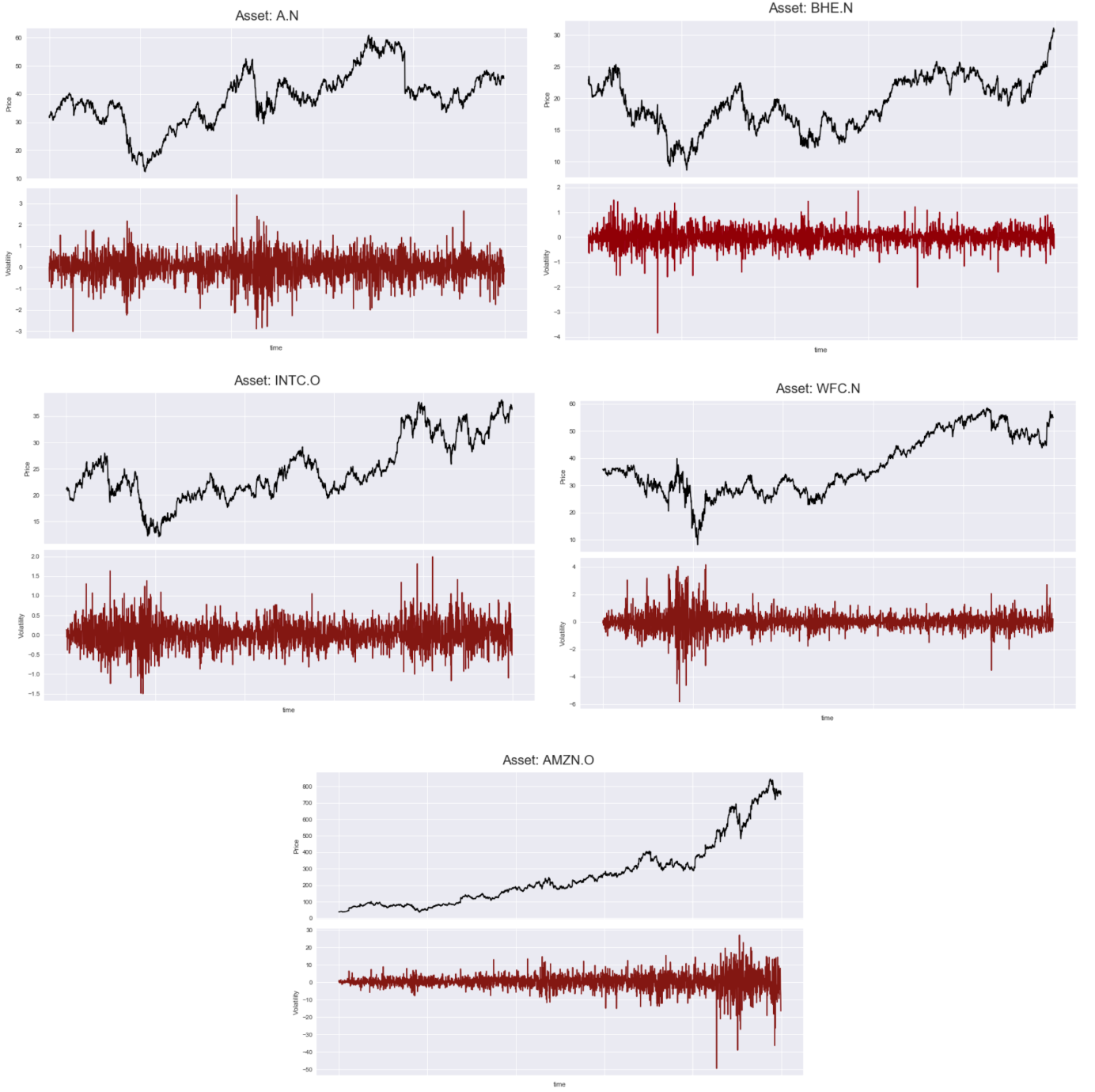}
    \caption{Price and volatility of small stocks (Agilent, Benchmark Electronics - top) and large stocks (Intel, Wells Fargo, Amazon - bottom)}
\end{figure}

Figure 2 clearly shows one of the main challenges of FTS forecasting which is the change of volatility over time. Indeed, looking at the volatility of Amazon, one notices a significant increase in volatility with increasing size of the company. This is the main justification behind picking a diverse set of stocks that are bound by different statistical properties.

It is worth showing some loss curves for the benchmark LSTM in order to examine how well the model generalises to the validation data with increasing numbers of epochs. This is shown in Figure 3 as follows.

\begin{figure}[!h]
	\includegraphics[width=195pt]{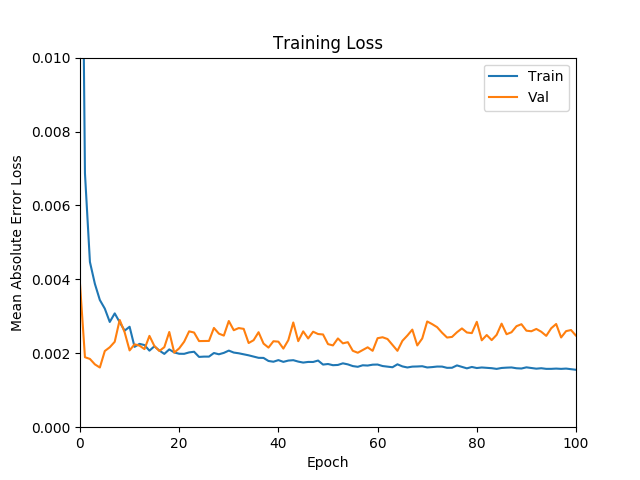}
	\includegraphics[width=195pt]{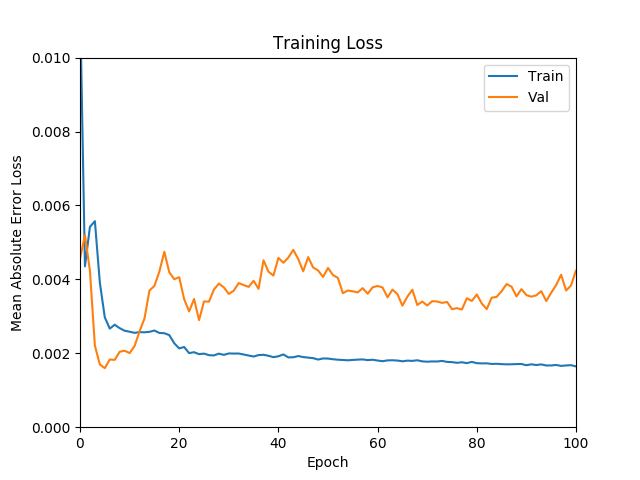}
	\caption{Typical LSTM loss for training and validation data per epoch}
\end{figure}

Figure 3 shows the typical LSTM training and validation losses through epochs. It can be noted that loss decreases as expected and begins to overfit after a certain number of epochs. It is worth noting one particularity from this plot which is that it shows training loss greater than validation loss for early epochs. While this may seem unusual, it is a documented artifact of using Keras. Indeed, the training loss output is the average loss over the batch while the validation loss is the final loss. It is therefore expected that the first few iterations show a higher training loss than validation loss. This is because the very first training iterations will have a higher loss than the final iteration of the validation loss. 

In order to tune this benchmark model, a grid search is undertaken. The LSTM performance during this grid search hyperparameter tuning, evaluated on the validation set, can be seen by the following 3D loss plots.

\begin{figure}[!h]
	\includegraphics[width=395pt]{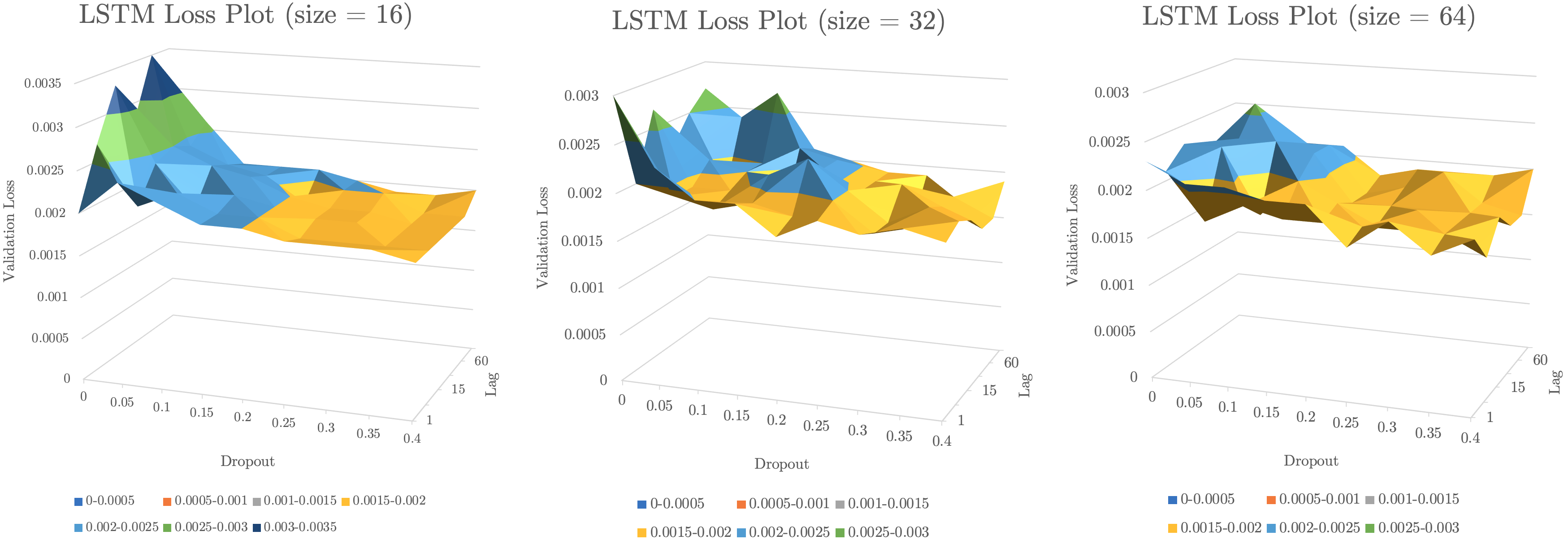}
	\caption{LSTM grid search loss plots for 3 different sizes}
\end{figure}

These plots in figure 4 show the performance impact of dropout and lag for three different LSTM sizes. Indeed, it seems the LSTM with size 64 performs best on average. In addition, dropout is clearly a useful regulariser for all three LSTM sizes, particularly for values of 0.2 and higher as it leads to lower losses. From this hyperparameter tuning, the best chosen configuration was with $size=64$, $lag=15$ and $dropout=0.1$. This resulted in a loss of 0.000805 and an up-down accuracy of 0.572 (or 57\%). Indeed, this consistently outperforms random guessing, which would have a performance of 50\%, and is in line with top of the range algorithms which usually achieve around 60\% up-down accuracy. 

Once the hyperparameters are tuned, a common FTS investigation worth pursuing is that of dataset shuffling techniques. In FTS, the choice of which piece of data to use as the validation set is not trivial. Indeed, there exist a myriad of ways of doing this which must be carefully considered before comparing this LSTM to an LSTM with attention. The three methods investigated in this paper are visualised in figure 5. 

\begin{figure}[!h]
	\begin{center}
    \includegraphics[width=395pt]{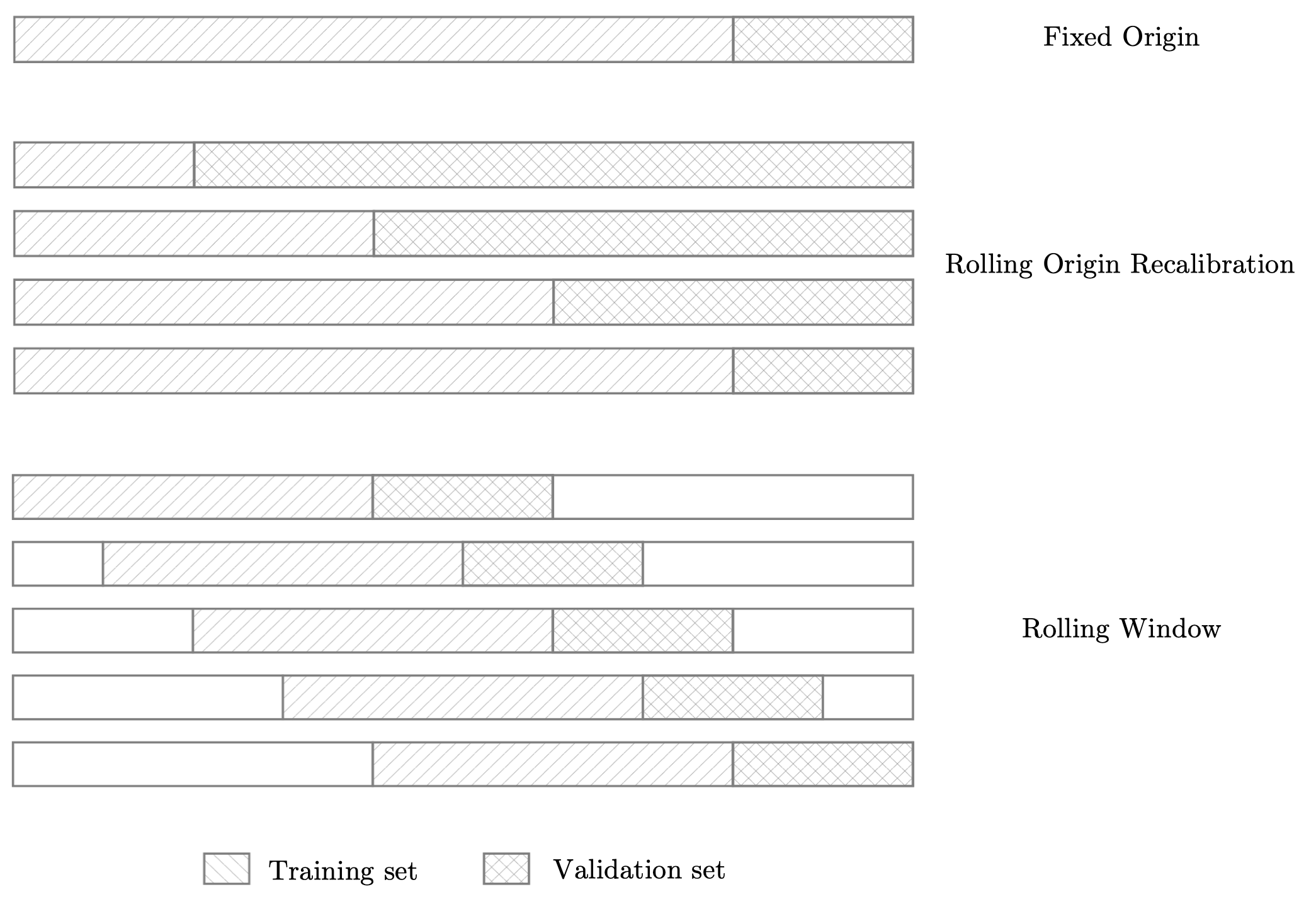}
	\end{center}
	\caption{Shuffling techniques visualised}
\end{figure}

The fixed origin method is the most naive and common method used. Given a certain split size, the start of the data is the training set and the end is the validation set. However, this is a particularly rudimentary method to choose, especially for a high-growth stock like Amazon. The reason why this is the case is that the Amazon's stock price starts off with low volatility and, as the stock grows, experiences increasingly volatile behaviour. We would therefore be training a model on low volatility dynamics and expect it to deal with unseen high volatility dynamics for its predictions. This has indeed shown itself to be difficult and come at a cost in performance for these types of stocks as we will see in table 1. Therefore our benchmark for validation loss and performance may be misleading if we only consider this. However, for stocks like Intel that are more constant in their volatility, this method is reasonable. 

The rolling origin recalibration method is slightly less vulnerable than fixed origin as it allows the validation loss to be computed by taking the average of various different splits of the data to avoid running into unrepresentative issues with high volatility timeframes. 

Finally, the rolling window method is usually one of the most useful methods as it is particularly used for FTS algorithms being run for long timeframes. Indeed, this model outputs the average validation error of multiple rolling windows of data. This means the final values we get are more representative of recent model performance, as we are less biased by strong or poor performance in the distant past.  

It is now important to show the model performance of our tuned benchmark LSTM using these different shuffling techniques. Using regular fixed origin our tuned LSTM achieved a loss of 0.000805 and an up-down accuracy of 57\%. Let's compare these values with top six values of rolling window (RW) and rolling origin recalibration (ROR) in table 1.

\begin{center}
 \begin{tabular}{||c c c c||} 
 \hline
 Loss (RW) & Accuracy (RW) & Loss (ROR) & Accuracy (ROR) \\ [0.5ex] 
 \hline\hline
 0.000692 & 0.538 & 0.000810 & 0.555 \\ 
 \hline
 0.000693 & 0.530 & 0.000825 & 0.571 \\
 \hline
 0.000725 & 0.575 & 0.000978 & 0.607 \\
 \hline
 0.000755 & 0.551 & 0.000989 & 0.563 \\
 \hline
 0.000780 & 0.514 & 0.001001 & 0.579 \\ 
 \hline
 0.000788 & 0.583 & 0.001014 & 0.538 \\  
 \hline
\end{tabular}
\end{center}
\begin{center}
    Table 1: Performance comparison of shuffling methods     
\end{center}

What table 1 shows is that both RW and ROR describe very slightly better performances (58\% and 60\%) than that of the simple fixed origin method. This suggests that for stocks like Amazon, using these shuffling methods would be inevitable.

Now that reasonable benchmarks have been ascertained, we can compare them to the performance of our LSTM with attention. The LSTM models with attention generated here had a strong tendency to overfit the training data for multiple epochs as shown in figure 6. 

\begin{figure}[!h]
	\includegraphics[width=195pt]{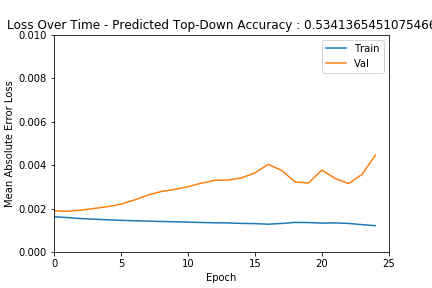}
	\includegraphics[width=195pt]{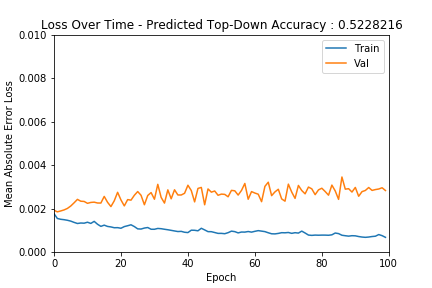}
	\caption{Typical LSTM+A loss for training and validation data per epoch}
\end{figure}

As shown in the loss curves of figure 6, we can see that performance at train time continually decreases while validation performance rises suggesting overfitting. Due to the high number of weights used in our implementation of the attention mechanism (with 16 cells, there are around 10 000 parameters), there are more parameters than the number of data points themselves, which inevitably leads to overfitting. Due to a large amount of parameters compared to the number of our data points, the loss values are very low and the training loss does not fluctuate a lot. This explains how the validation loss seems to increase immediately after one or two epochs when looking at the graphs. From this behaviour, we decided to limit the number of epochs for the LSTMs with attention (grid search over 25 epochs, compared to 100 epochs with LSTMs only). In order to help alleviate some of the overfitting errors and lack of generalisation, a hyperparameter grid search was undertaken to tune the model’s hyperparameters. In particular, dropout and varying the amount of lag for given LSTM sizes were investigated, as done for the LSTM without attention. Loss plots are shown in figure 7 below. 

\begin{figure}[!h]
	\includegraphics[width=395pt]{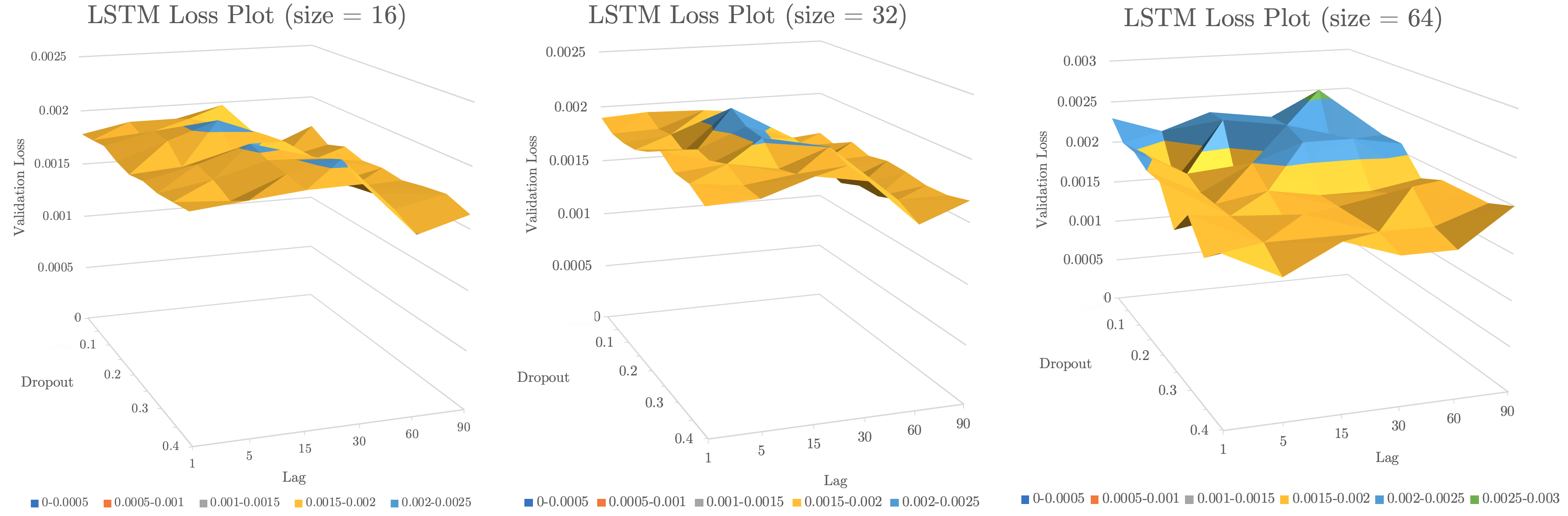}
	\caption{LSTM+A grid search loss plots for 3 different sizes}
\end{figure}

The LSTM with attention achieves a globally lower loss than the LSTM without attention but is sensitive to different hyperparameters. Indeed, in the case of the LSTM with attention, the amount of lag is a significant factor in decreasing loss. This can be seen in figure 7 as lags larger than 30 drastically improve loss. It could be hypothesized that this is due to the unique nature of attention that can focus on relevant information from the past while the LSTM without attention is incapable of doing this due to its long term dependencies. This seems to support our original hypothesis on the impact of using attention on LSTMs for FTS. From this hyperparameter tuning, the best chosen configuration was with $size=16$, $lag=60$ and $dropout=0.05$. This resulted in a loss of 0.001511 and an up-down accuracy of 0.588 (or 59\%). Again, this is in line with top of the range algorithms and is slightly higher than that of the LSTM without attention (58\% accuracy). 

It is now worth comparing the performance, loss and accuracy, of both the tuned LSTM and the tuned LSTM with attention across all five stocks. The optimal hyperparameters for both the LSTM and LSTM with attention were set as detailed previously and the following performances were observed.

\begin{center}
 \begin{tabular}{||c c c c c c||} 
 \hline
   & Intel & Wells Fargo & Amazon & Agilent & BE \\ [0.5ex] 
 \hline\hline
 LSTM (loss) & 0.000805 & 0.001200 & 0.002804 & 0.001073 & 0.002300 \\ 
 \hline
 LSTM (accuracy) & 0.573 & 0.457 & 0.490 & 0.457 & 0.470 \\
 \hline
 LSTM+A (loss) & 0.001511 & 0.000357 & 0.003168 & 0.000955 & 0.001787\\ 
 \hline
 LSTM+A (accuracy) & 0.588 & 0.603 & 0.328 & 0.443 & 0.493\\
 \hline
\end{tabular}
\end{center}
\begin{center}
    Table 2: LSTM and LSTM with attention performance comparison    
\end{center}

From table 2 above, we can observe multiple interesting dynamics at play. The first being that the optimal parameters chosen by grid search with the Intel stock fail to generalise to other stocks for the LSTM and the LSTM with attention. While this was somewhat expected it is interesting nonetheless to observe this limitation. Further time would allow the tuning of hyperparameters for each stock but this is outside the scope of this paper and will be discussed in the following section. In addition, it seems that the LSTM with attention has a higher variability in its performances. This makes sense as we know that the model is very complex in terms of parameter number with respect to the data input. Overall in this comparison, the LSTM with attention has outperformed the regular LSTM. Indeed, it does seem like we can tentatively confirm our original hypothesis albeit further work is certainly required. In addition, the attention architecture used in this paper does involve certain limitations and has a caveat of inherent complexity. 

\section{Limitations and Future Work}

One particular limitation of this model is that we only considered the first prediction in our accuracy score. This means we are effectively doing single step ahead prediction even if the model is inherently capable of doing multi-step ahead prediction. Indeed, multi-step ahead forecasting is historically a much harder problem than single step ahead. The reason why this is the case can be seen by the naive approach of converting single step ahead forecasting to multi-step forecasting via the iterative method. The iterative method simply takes the predicted output of the model for time $t+1$ and forecasts output for $t+2$ from the existing forecast without feedback from the real world. This causes the model error to cascade through iterations and grow out of control for more than a handful of time steps. This is a fundamental issue where most models fail for FTS forecasting. However, the model described in this paper is capable of seq2seq which allows for forecasting an entire sequence without the uncontrollable error growth experienced by iterative methods. Indeed, this potential for seq2seq is key advantage of the models used and constitutes a major avenue for future work and research. 

Another limitation of this investigation is confidence interval encoding. Indeed, there are various ways to do so but the approach used here was to make the models output a value between -1 and 1 by using a $\tanh$ function. However, the confidence output of our model was not considered in the up-down accuracy detailed above. Indeed, a simple way to exploit confidence intervals to improve performance would be to only execute trades beyond certain threshold of confidence. Indeed, this has been leveraged successfully for most models in FTS forecasting however it remains a non-trivial task. Indeed, the higher the threshold is set for executing trades, the higher the overall up-down accuracy will be. However, the higher the threshold the lower the number of total trades executed. There is therefore a major trade-off here as perfect accuracy for few trades may be less profitable than excellent accuracy for many trades.  

One major possible extension to the work presented here would be the use of Bayesian optimisation. Bayesian optimisation is a technique commonly used in financial time series forecasting for the tuning of hyperparameters. While this was not done in this report for time constraints, it is essential for a more detailed comparison of the potential of attention mechanisms in LSTMs for FTS forecasting. Indeed, as revealed by this paper, hyperparameters have different sensitivities in the benchmark LSTM and in the LSTM with attention. Future work would consist of choosing an appropriate surrogate function and acquisition function (such as expected improvement) in an attempt to explore the hyperparameter space with a smart explore-exploit trade-off. This is indeed still an open problem in the field of FTS forecasting.

Another possible extension would be to investigate how the LSTM with attention performs compared to other benchmark models for FTS forecasting. One particular interesting comparison could be done with cutting edge Temporal Convolutional Networks (TCN) as they are currently showing promising performances compared to RNNs in NLP applications.

\section{Conclusions}

This paper has demonstrated the performance of a benchmark LSTM. Investigations were undertaken to avoid using a single number as a benchmark and a wide range of experimentation was conducted. This includes the investigation of data shuffling methods such as rolling window and rolling origin recalibration, which showed the impact of volatility on estimating model performance. Ultimately, the benchmark LSTM consistently performed around the 58\% range which is in line with the best models currently available which usually reach performances around 60\%. 

In addition, the LSTM with attention was successfully implemented and leveraged for FTS forecasting. This task, while totally novel, did nonetheless closely follow methods used in other FTS papers and papers using attention for NLP. This LSTM with attention achieved performances of around 60\% and above, albeit with a higher variability than the benchmark LSTM.

The final comparison of this LSTM with attention does indeed confirm the investigated hypothesis that an LSTM with attention can improve the performance of existing LSTMs in FTS. A slight improvement was indeed highlighted in the final comparison table, albeit both models need to be re-tuned in between stocks. A theoretical explanation for why this is the case was suggested, developed and tested. 

Finally, further work in this topic has been suggested and the main model limitations were discussed.


\bibliographystyle{plain}
\bibliography{ref}

\end{document}